\pdfoutput=1

\documentclass[11pt]{article}

\usepackage[]{ACL2024}

\usepackage{times}
\usepackage{latexsym}

\usepackage[T1]{fontenc}

\usepackage[utf8]{inputenc}

\usepackage{microtype}

\usepackage{inconsolata}

\usepackage{graphicx}
\graphicspath{{graphs/}}
\usepackage{amsmath}
\usepackage{physics}
\usepackage{subcaption}
\newcommand{\helmcite}{\citep{Hendrycks2020MeasuringMM,clark2019boolq,kocisky-etal-2018-narrativeqa,Zellers2019HellaSwagCA,Mihaylov2018CanAS,kwiatkowski-etal-2019-natural,maas2011imdb,Campos2016MSMA,narayan-etal-2018-dont,lin-etal-2022-truthfulqa,see-etal-2017-get,choi-etal-2018-quac}}
\title{Efficient Benchmarking (of Language Models) }

\author{\normalsize
Yotam Perlitz\qquad Elron Bandel \qquad Ariel Gera \qquad Ofir Arviv \\ 
\textbf{ \normalsize Liat Ein-Dor} \quad \textbf{ \normalsize Eyal Shnarch} \quad \textbf{ \normalsize Michal Shmueli-Scheuer} \quad \textbf{ \normalsize Leshem Choshen} 
\\
IBM Research AI \\
\texttt{\{yotam.perlitz,leshem.choshen\}@ibm.com}}

\usepackage{amsmath,amssymb}
\usepackage{algorithm}
\usepackage{algpseudocode}
\usepackage{xfrac}
\usepackage{dsfont}
\usepackage{tcolorbox}
\usepackage{subcaption}
\usepackage{enumitem}
\usepackage{newfloat}
\newcommand{\slack}[1]{}
\newcommand{\cready}[1]{}
\newcommand{\anonm}[1]{}
\DeclareMathOperator*{\EX}{\mathbb{E}}

\newcommand{\stabilitymeasure}{\emph{DIoR}}

\newcommand{\stabilitymeasurelong}{Decision Impact on Reliability}

\newcommand{\helm}{HELM}
\newcommand{\flashhelm}{\emph{Flash-HELM}}

\begin{document}
\maketitle
\begin{abstract}
The increasing versatility of language models (LMs) has given rise to a new class of benchmarks that comprehensively assess a broad range of capabilities. 
Such benchmarks are associated with massive computational costs, extending to thousands of GPU hours per model. However, the efficiency aspect of these evaluation efforts had raised little discussion in the literature.

In this work, we present the problem of \emph{Efficient Benchmarking}, namely, intelligently reducing the computation costs of LM evaluation without compromising \emph{reliability}. 
Using the \helm{} benchmark as a test case, we investigate how different benchmark design choices affect the computation-reliability trade-off. 
We propose to evaluate the reliability of such decisions, by using a new measure -- \stabilitymeasurelong{},  \stabilitymeasure{} for short.
We find, for example, that a benchmark leader may change by merely removing a low-ranked model from the benchmark, and observe that a correct benchmark ranking can be obtained by considering only a fraction of the evaluation examples.
Based on our findings, we outline a set of concrete recommendations for efficient benchmark design and utilization practices. To take a step further, we use our findings to propose an evaluation algorithm, that, when applied to the HELM benchmark, leads to dramatic cost savings with minimal loss of benchmark reliability, often reducing computation by x100 or more.
\end{abstract}

\section{Introduction} \label{sec:intro}
\begin{figure}[t]
\centering
\includegraphics[width=\columnwidth]{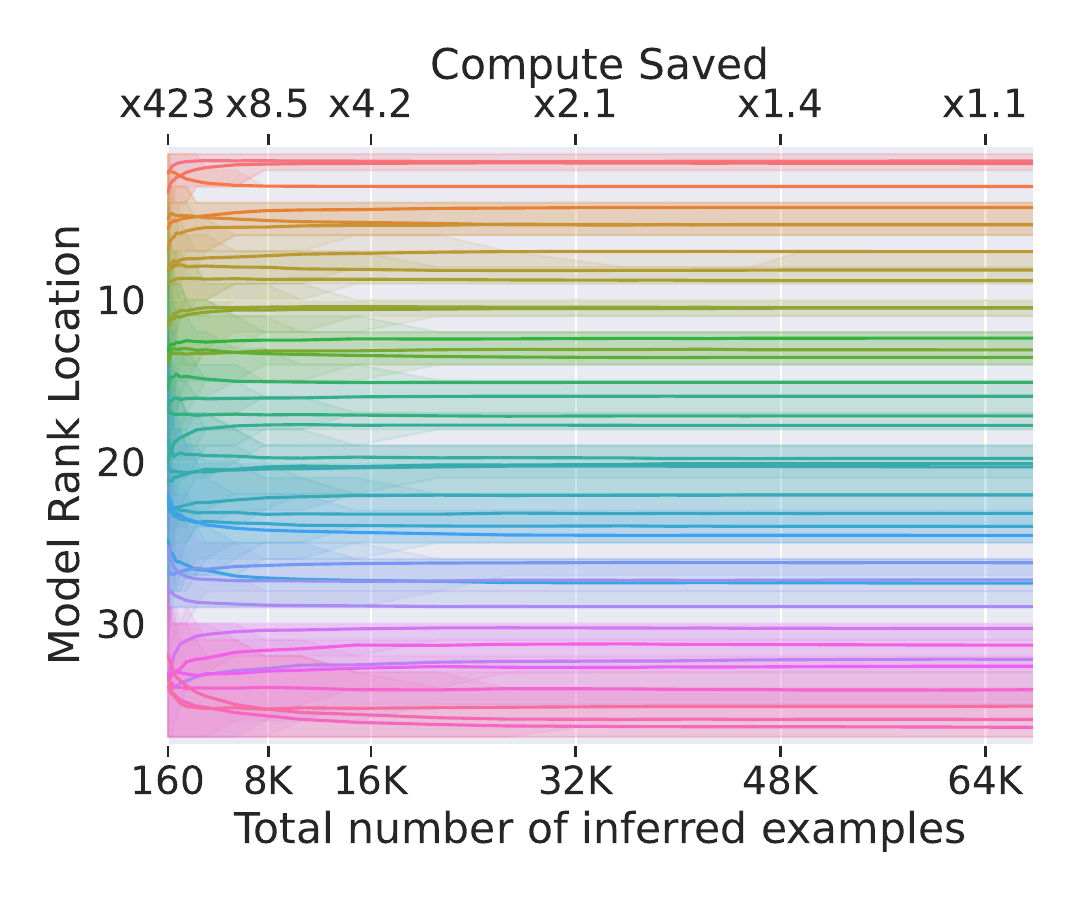}
\caption{\textbf{\helm{} model ranks for different numbers of inference calls.} Each colored curve represents a different model. Model ranks are extremely stable even when compute drops dramatically: a $\times10$ decrease in the number of examples per scenario produces nearly the same results as the full benchmark, while a $\times400$ reduction still clusters models in the same small groups seen in the full compute regime.}
\label{fig:rank}
\end{figure}

Given the ongoing advances in the versatility and performance of Language Models (LMs), they are now expected to perform a diverse range of tasks. 
This expectation raises a profound challenge -- how do we evaluate and rank the quality of different LMs over a variety of capabilities?

This is a complex evaluation endeavor \citep{chang2023survey}, as it transcends the boundaries of a specific task and seeks to measure the overall capabilities of an LM over a wide manifold of natural language tasks. 
To this end, LM benchmarks are constantly being proposed, where each new benchmark further expands the coverage and diversity of evaluated tasks and settings \citep{wang-etal-2018-glue,bigbench,lmeval-harness,talmor-etal-2020-olmpics,yuan2023evaluating,zhang2023dialogstudio}. 
Running such expansive benchmarks can entail spending \$$10$K+ or $4$K+ GPU hours for evaluating a single model \citep{liang2022holistic}, and may even surpass those of pretraining \citep{biderman2023emergent} when evaluating checkpoints.
At the same time, even when compute resources are abundant, benchmarks are bound to make certain concessions aiming to \emph{approximate} true model ability. 
These concessions --  in the form of benchmark design choices -- are to be made such that their impact on benchmark reliability (\S\ref{sec:reliability}) is both minimized and transparent. This, to minimize cases where sub-optimal design choices lead to reliability issues such as anointing a different best model or making rank differences between models statistically meaningless. 

In this work, we call attention to the topic of \textbf{Efficient Benchmarking}, namely intelligently reducing the computation costs of evaluation without compromising reliability. 
While the trade-off between computation and performance is usually discussed in the context of pre-training (e.g., scaling laws;  \citealp{hoffmann2022trainingChinchilla,ivgi2022scaling}) and fine-tuning (e.g., parameter efficient; \citealp{lialin2023scaling}), here we call for putting this trade-off on the center stage of evaluation design. 

In practice, the compute side of the trade-off already plays a role in most large-scale evaluation decisions, both in benchmark design \citep{liang2022holistic} and in its use for evaluation \citep[e.g., choosing the number of seeds;][]{csordas2021devil, choshen2022start}.
However, despite their practical importance, these choices and their impact on benchmark reliability have hardly been discussed in the literature, making researchers apply their own efficiency heuristics instead of using systematic guidelines or literature when building their benchmarks. 

In order to advance efficient evaluation practices, the community is in need of a systematic set of guidelines and recommendations. 
These, in turn, must be based on a rigorous study of the different decisions made in benchmark design and how they affect efficiency. 

To begin addressing these challenges, we propose \stabilitymeasurelong{} -- \stabilitymeasure{} -- a way to measure the Impact of a Decision over a setup size (e.g., $1$K examples, $10$ datasets) on the Reliability. In addition, we perform a comprehensive analysis study on efficient benchmarking. 
With \helm{} \citep{liang2022holistic} as a test case, we test various decisions made and how they affect the trade-off between computation and reliability:
decisions about scenarios which are aggregated phenomena (\S\ref{sec:scenario}), subscenarios (\S\ref{sec:subscenarios}), few-shot prompts (\S\ref{sec:seeds}) and the metric (\S\ref{sec:metric}). 
Among other findings, 
we observe a substantial computation redundancy (see Fig.~\ref{fig:rank}, \ref{fig:seeds}); 
that a change in one rank is currently unreliable (\S\ref{sec:scenario}); 
that splitting the data into groups (scenarios) hurts reliability; 
and that the mean win rate score (\S\ref{sec:metric}) is unreliable and gameable. 

Given our analysis findings, we collect a set of general guidelines for future benchmark creation and use (see Tips above). Moreover, we show how our findings can benefit current benchmarks by proposing \flashhelm{} (\S\ref{sec:flashHELM}), a general evaluation algorithm that enables obtaining a model's ranking with a fraction of the computation and minimal loss of benchmark reliability.

\definecolor{main}{HTML}{13274F}

\begin{figure*}[t] \label{tipsbox}
\begin{tcolorbox}[width=\textwidth, colback=main!10, colframe=main, center title, fonttitle=\large\bfseries, title=\color{white} 
 \textbf{Efficient Benchmark Building Checklist} ]

\textbf{\checkmark \hspace{0.2em} Report Benchmark compute costs  (\S\ref{sec:intro})} 

Benchmarks often have heavy compute requirements, report required compute to increase usability.\vspace{1ex}

\textbf{\checkmark \hspace{0.2em} Verify your design decisions with \stabilitymeasure{} (\S\ref{sec:reliability})}

Quantify your benchmark's reliability-compute trade-off across your different decisions. For example: did you use enough examples/scenarios/prompts/seeds? perhaps too many?
 \vspace{1ex}

\textbf{\checkmark \hspace{0.2em}  Compute matters - Suggest an Efficient benchmark version (\S\ref{sec:examples} and \S\ref{sec:flashHELM})}

In addition to the full benchmark, provide the user with efficient compute-saving alternatives with varying degrees of reliability, e.g., by reducing the number of examples.
 \vspace{1ex}

\textbf{\checkmark \hspace{0.2em}  Reliability matters - Report where it is lacking (\S\ref{sec:examples})}

Identify reliability issues, such as distinguishing between top models for which we found HELM to be unreliable. Transparently report these limitations to avoid over-interpreting unreliable results.\vspace{1ex}

\textbf{\checkmark \hspace{0.2em} Maximize data-points variability to improve reliability (\S\ref{sec:seeds})}

When sampling from multiple sources of variation (e.g., prompts, examples), maximize the coverage of each source, rather than exhausting all cross-product combinations of a few sources.
 \vspace{1ex}

\textbf{\checkmark \hspace{0.2em}  Don't aggregate if possible, it hurts reliability (\S\ref{sec:subscenarios})}

When possible, avoid aggregating scores from distinct phenomena into a single metric, this will reduce the reliability of the overall benchmark score. Keep scores disaggregated when meaningful.

\end{tcolorbox}
\end{figure*}

In summary, the contributions of this work are as follows:
\begin{enumerate}
    \item We highlight the importance of the \textbf{balance between computation and reliability} in benchmark design and utilization, and propose \textit{\stabilitymeasure{}} as a quantitative measure of the reliability of a specific efficiency strategy.
    \item We conduct the first \textbf{systematic study} of the effects of benchmark design on reliability.
    \item Given the analysis findings, we provide an \textit{Efficient benchmark building checklist}; These guidelines outline how best to reduce the computational cost of benchmarking while maintaining an adequate level of evaluation reliability.
    \item We propose an \textbf{algorithm for dynamic ranking} of a new LM, assigning higher importance to rank top-performing models. In \helm{}, we show that this algorithm (later incorporated into the framework\footnote{\url{https://crfm-helm.readthedocs.io/en/latest/get_helm_rank/}}) dramatically reduces the computation by up to $\times200$ with minor deviations from the original ranking (see Fig.~\ref{fig:tournament}).
\end{enumerate}

\section{The Objective, Validity, Reliability} \label{sec:reliability}
In this section, we first define 3 critical aspects for evaluation: \emph{the objective}, \emph{validity}, and \emph{reliability}. 
Then, we discuss benchmark reliability and how to measure it, in more detail, being the focus of this study.

\paragraph{The Objective.} 
The question the benchmark aims to answer. 
For example, ``How good is a given model at sentiment analysis?'' or ``Which is the best language understanding model?''. The objective guides the initial, high-level decisions such as the choice of metric, tasks, domains, and datasets. 
\paragraph{Validity.} 
Ensuring that the benchmark actually satisfies the objective, i.e., that it answers the right question, is not trivial.
Following common psychometrics literature \citep{cronbach1946response}, we refer to this quality as \emph{validity}. 
Validity challenges are often discussed in the literature, in general, \citep{v2015build} and in validating metrics \citep{choshen-abend-2018-automatic,freitag2022results} or data \citep{poliak2018hypothesis,gururangan2018annotation,northcutt2021confident,northcutt2021pervasive}.
For example, if the objective is to measure broad language understanding capabilities but the benchmark measures only a narrow aspect of language understanding, the benchmark has a validity problem.

\paragraph{Reliability.} Due to the noisy nature of broad evaluation, two valid protocols may yield different results \citep{maynez-etal-2023-benchmarking}. \emph{Reliability} assesses the degree to which the evaluation answer remains consistent under different random decisions, many of which are selections from the distribution of elements composing the benchmark \citep{kuder1937theory}.


Building a benchmark involves numerous decisions (e.g., the number of datasets, or of examples per dataset). Importantly, such design decisions determine the reliability of the benchmark, and the conclusions that can (or cannot) be drawn from it.
Therefore, we argue that such decisions must be made in an informed manner, including considering their impact on reliability.
From a practical point-of-view, well-informed decisions can lead to improved benchmarks, yielding more reliable results with lower computational costs.

\subsection{Quantifying Reliability: \stabilitymeasure{}}\label{subsec:DIOR}
Just like significance, which relies on p-value, reliability requires a metric. However, such a metric is not available. Thus, we propose a new metric -- the \stabilitymeasurelong{} test (hereafter \stabilitymeasure{}) -- as a way to assess the effect of a benchmark design decision (e.g., choosing $16$ specific language understanding datasets) on the reliability of the benchmark. 
Given a collection of instantiations of the decision (e.g., dataset samples of size $16$), a benchmark scoring function (e.g., rank) and a similarity meta-metric to measure the consistency of the scoring function under a pair of different instantiations (e.g., correlation between rankings), \stabilitymeasure{} assesses the stability of the meta-metric across different instantiations. Specifically, we define \stabilitymeasure{} as the lower bound of the confidence interval for the value of the meta-metric; we report the lower bound as this corresponds to the minimal value we are certain of.

Formally, given a set of models $M$, random instantiations of the decision $c\sim D$, and the original instantiation $c_o$, a benchmark scoring function $s_c\colon M\to r$ and a similarity meta-metric $f \colon r,r \to [0,1]$, \stabilitymeasure{} is defined as:
$$\textsc{DIoR}=CI_{95\%,c\sim D}(f(s_{c_o}(M),s_{c}(M)))$$ 

A reliable decision should receive a high \stabilitymeasure{}, implying that different instantiations do not substantially affect the results.

\section{Data, Models and Scores}\label{sec:helm_details}
As a test case for investigating benchmark efficiency, we analyze the results of the \helm{} benchmark \citep{liang2022holistic}. We stress, that although \helm{} satisfies a good candidate for our analysis, due to the wide range of tasks and models it offers, our conclusions and methods are general and in no way bound to a specific benchmark.

We take the scores of $37$ models reported on \helm{} version 0.2.2\footnote{\url{https://crfm.stanford.edu/helm/v0.2.2/?group=core_scenarios}} as the data for most of our experiments. As a test set for our recommendations (\S\ref{sec:flashHELM}), we take the $7$ new models introduced in the latest the later version, 0.2.3.


The \helm{} benchmark defines a taxonomy of \textit{scenarios}, where each scenario corresponds to a collection of labeled data, as well as a metric used to evaluate performance on this data. The benchmark designates $16$ scenarios~\helmcite{} 
as \textit{``core scenarios''}, on which all LMs are evaluated and a bottom-line score is calculated (see below).
Each scenario within \helm{} is further divided into one or more \textit{subscenarios}; each is an individual dataset with a dedicated scoring function and 3 few-shot prompts, which are originally referred to as seeds. 
Note that the grouping of subscenarios into a scenario can stem from historical reasons, such as grouping datasets based on prior work. For instance, one of the scenarios in \helm{} is \textit{RAFT}, which consists of several different datasets used within the RAFT benchmark \citep{alex2021raft}.
In \S\ref{sec:subscenarios}, we discuss the consequences of this grouping decision. 

The \helm{} benchmark ranks LMs by an aggregation of their scores over all $16$ core scenarios and 65K examples. The aggregation metric used is \textit{Mean Win Rate (MWR)}, which compares LMs against one another per scenario \citep[a Borda Count variant, ][]{emerson2013original}. MWR measures the average win rate for each model over all scenarios (see App.~\ref{ap:mwr_def} for a formal definition). 

\section{Experimental Setting}
In our main experiments, we calculate \stabilitymeasure{} to examine the reliability under the current realization of the benchmark, as well as more efficient realizations. 
Thus, we calculate \stabilitymeasure{} for varying amounts of compute, ranging from the full \helm{} benchmark to a small fraction of it (e.g., a benchmark with $1$ scenario, or $100$ examples). 

For each design choice (number of examples, scenarios etc.) we sample different instantiations of this choice, and use them to calculate \stabilitymeasure{}. We follow a bootstrap approach, namely, sampling $1$K times with repetition.
For example (in \S\ref{sec:scenario}), to test whether taking $10$ scenarios reliably indicates the best model, we sample $10$ scenarios (out of the available $16$) $1$K times, calculating the win rate values for each sample (in a sample, some datasets may be chosen more than once, or not at all).


\subsection{Benchmark Objectives}
Throughout our analysis, we consider three objectives that benchmarks often aim to measure. For every objective, we recommend a specific metric and then provide a related meta-metric to check its reliability. 

One objective is to acquire the \textbf{full ranking}. 
The meta-metric measures the number of models switching places in the overall ranking (Kendall $\tau$). We also calculate a weighted alternative that emphasizes correctly ranking the top models \citep{Vigna2014AWC}, finding generally similar trends (see App.~\S\ref{ap:obj}, \S\ref{ap:full_subscenario}). 

For the objective of determining which model is the \textbf{best model}, we define the meta-metric as the \emph{Error Rate}, namely, the probability (across different instantiations) of a rank switch between the top two models.
As we care about the best model in general, and not the current one specifically, we repeat the experiment $5$ times, each time removing the top model from the benchmark (as if it was not yet submitted). 

The last objective is to evaluate \textbf{model quality}, i.e., how well each model performs. For this, we calculate the absolute bottom-line score. To be consistent with the literature, we report MWR as the model quality metric, where the meta-metric is the absolute difference in MWR scores.

\section{Results}

In this section, we examine the impact of different design choices on the reliability of the benchmark objectives.

\subsection{Scenarios}\label{sec:scenario}
\helm{} selected $16$ core scenarios for model evaluation. 
We do not challenge this choice's validity or relevance.
Instead, by applying bootstrapping, we run a simulation of selecting equally valid alternative scenarios, in order to investigate the reliability of this choice.

\begin{figure}
\centering
\includegraphics[width=\columnwidth]{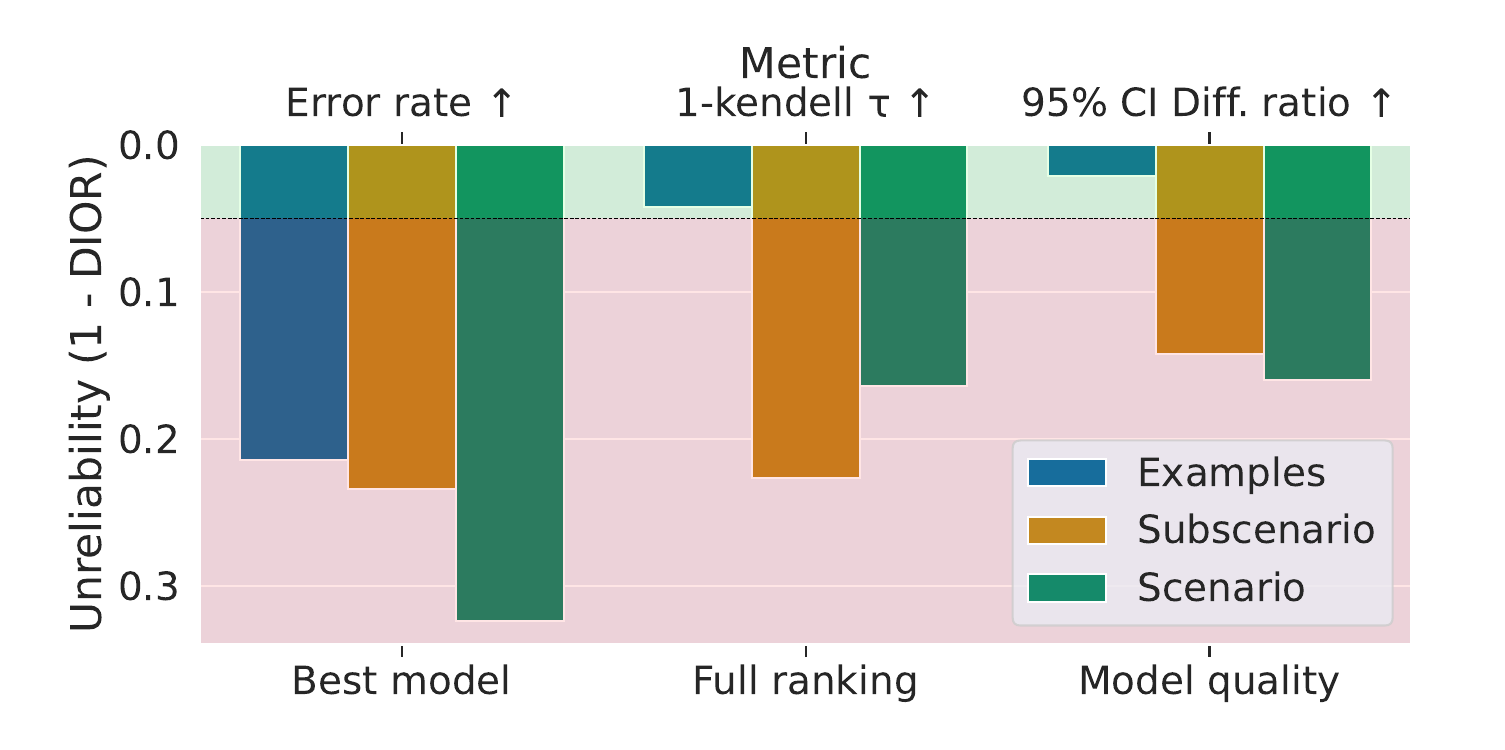}
\caption{\textbf{Scenarios / Subscenarios / Examples \stabilitymeasure{}.} Different subscenarios or scenarios would highly affect results, but not examples. Each cluster of bars represents a measure of \stabilitymeasure{} (top labels) for the corresponding objective (bottom labels). Each color denotes the granularity: Examples, Subscenarios, or Scenarios. The area above the vertical line in light green represents high reliability levels ($\ge 95\%$), while the area below in red indicates lower reliability.
\label{fig:boot_compare}}
\end{figure}

In Fig.~\ref{fig:boot_compare} we report the reliability of \helm{}'s original choice of ($16$) scenarios, for each of the objectives. 
We find that the reliability of the set of scenarios is low. 
Put another way,
under a different choice of scenarios it is quite likely that \helm{}'s ranking, score, and winners would be different.
Further, as shown in App.~\S\ref{ap:obj}, upon reducing the number of scenarios, reliability drops drastically; 
thus, the common compute-reduction approach of dropping datasets (e.g., big-bench lite, \citealp{bigbench}), is in fact an ill-advised practice.

\subsection{Subscenarios} \label{sec:subscenarios}

In this section, we first examine the reliability of \helm{}'s original design choice of $40$ subscenarios (single-datasets grouped to construct the scenarios).  
Then, we question the reliability of the common design choice (also prevalent in \helm{}) of grouping multiple subscenarios into scenarios vs. keeping every subscenario as a standalone. 

Repeating the reliability test for the three objectives (Fig.~\ref{fig:boot_compare} and App.~\S\ref{ap:obj}), we find that similarly to scenarios, the choice of subscenarios only supports low reliability, meaning that dropping subscenarios is a problematic approach for reducing compute.
Given this finding, we revisit the decision to group subscenarios into scenarios. 
We find, that in terms of reliability, considering each subscenario as a standalone scenario is helpful, for example, in reducing the error rate between top pairs to 14\% instead of 22\% (see App.~\ref{ap:scenarios_vs_subscenarios}). 

The rationale for grouping subscenarios is their shared focus on testing a particular skill or phenomenon. 
Their reweighting as a single group prevents over-emphasis on this skill across the benchmark (see example in App.~\S\ref{ap:examples}).
By complementing each other, these grouped subscenarios should offer a more holistic assessment of that specific skill. 
Qualitatively, in \helm{}, it is not clear that the grouping is crucial and indeed prevents over-representation of specific phenomena. 
For example, the 7 open/closed question answering \textit{scenarios} (e.g., openbookQA, \citealp{Mihaylov2018CanAS}) seem closer in spirit to each other, than MMLU's \citep{Hendrycks2020MeasuringMM} $4$ \textit{subscenarios} which were designed to cover distinct topics in language understanding.

If the above intuition proves correct, related subscenarios should test the same skill and are expected to rank models consistently. Conversely, rankings from unrelated subscenarios would likely diverge, as they evaluate different capabilities.

In App.~\S\ref{ap:full_subscenario}, we measure just that and present the correlation between rankings made by different subscenarios. 
We do not find an a stronger correlation within subscenarios that belong to the same scenarios, concluding that aggregation is not needed for \emph{validity}. 

Although we found that aggregating subscenarios scores hurts the general benchmark reliability, we note that aggregated scores might still be interesting to report for fine-grained evaluation \citep{gehrmann2021gem} or for historical reasons (reusing a benchmark) as these can be considered as separate sub-objectives. 
Hence, we suggest that each such sub-objective would include aggregations, but that the bottom-line benchmark calculations should ignore these. This will allow the benefits of sub-objectives while preventing overall benchmark reliability decline. 
Concretely, in \helm{}, one can aggregate the final model score over all \textit{subscenarios}, but still report the aggregated scores per \textit{scenario} separately.


\subsection{Examples}\label{sec:examples}
Previously, we have found that the reliability of (sub)scenarios is already low, hence decreasing computational cost by removing them is undesirable. In contrast, as Fig.~\ref{fig:boot_compare} shows, the current choice of examples is highly reliable. Thus, removing examples is a preferable strategy for reducing compute. 
Further, we find a certain discrepancy between the objectives, where best-model is not reliable while model-quality and full-ranking are. 
To reiterate, in the current state of HELM, discussing the top model is pointless. In smaller benchmarks, we expect the problem to be even more severe.

\begin{figure}[t]
\centering
\includegraphics[width=\columnwidth]{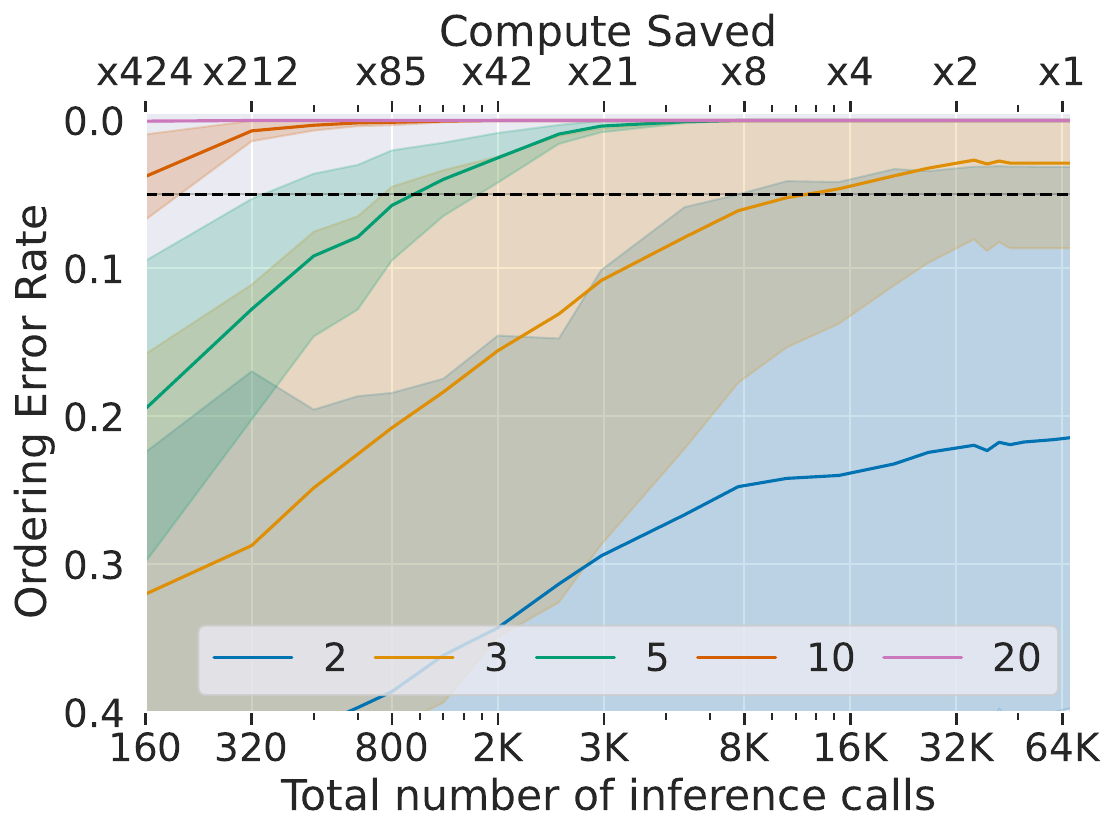}
\caption{\textbf{The probability that models would switch places} (y-axis) given a different random choice of examples for evaluation (x-axis). Each line corresponds to taking a group of N models and testing if the top and bottom switch places. Results are averaged across $1$K iterations ($95$\% confidence interval in shade) and over the top $5$ models as the top model.}
\label{fig:subset_pairsig}
\end{figure}

For the model-quality and full-ranking objective, the current state is reliable, hence, we test reliability with fewer examples per scenario.
We find (see Fig.~\ref{fig:rank}) that model ranks are quite stable regardless of the number of examples used. 
Remarkably, with the bare minimum of examples, models are already clustered into equivalence classes of about $2$-$5$ models, and with a few hundred examples, models are separated into groups of ${\sim}2$ -- the best separation the benchmark ever achieves. 
We find similar trends, of high reliability with a small number of examples, for the other objectives as well (App.~\S\ref{ap:obj}). Fig.~\ref{fig:locs_res} quantifies the error in rank per model, and finds it is small, ranging from $6$ to $2$.

In the findings discussed so far, we repeatedly found models to be indistinguishable from the model ranked right above or below them. 
However, it is also interesting to consider the level of separation between models that are farther apart. 
Thus, we examine clusters of adjacent models within the full HELM ranking (e.g., for a cluster size of $5$, we consider the models ranked $1$-$5$, $2$-$6$, $3$-$7$ etc.). 
In Fig.~\ref{fig:subset_pairsig} we plot the probability of rank location switch (i.e., error rate) between the first and last models in rank clusters of sizes $2$, $3$, $5$, $10$ or $20$.
While with cluster size $2$ models often switch places -- even with all benchmark examples -- for clusters of $3$ (i.e., a diff of two places), \sfrac{1}{4} of the computation is sufficient to get an average error rate under 5\%. For larger clusters, one can get reliable results with a hundredth of the cost or less.

We leave the special case where a ``benchmark'' is a single dataset to App.~\S\ref{ap:standalone}. Even then, fewer examples suffice. Moreover, as some datasets are more stable than others, one can tune the number of examples per dataset as needed, taking more examples where distinctions are harder to make. We leave more elaborate research on that for future work.

\subsection{Few-Shot Prompts (\helm{}'s seeds)}\label{sec:seeds}
Under the in-context learning paradigm, LMs are expected to predict the right answer given some examples. 
As the choice of exemplars  might change results \citep{min2022rethinking, dai2023can,pan2023context}, a reliable benchmark should account for this variability as well. 
For this reason, \helm{} considers three sets of few-shot exemplars, uses them against every example \citep{liang2022holistic}, and averages their score. 

To assess the reliability of prompts, bootstrap approximation is not a viable option as there are only three prompts.
Instead, we compare the effect of two different approaches for using a given budget to evaluate model performance. 
In our example, the budget of inference calls is $3$K examples. 
One method, as \helm{} did, samples a set of K examples and then samples prompts ($3$). Then, every model is tested on every example and prompt, the full cross-product. 
In contrast, one may sample \emph{uniformly} from the cross product of all (K) examples and all possible prompts, where each call samples a different prompt and example (and perhaps other traits), ideally a unique example and prompt in each call. 
Thus the calls will evaluate $3$K examples and $3$K prompts. 
This approach captures more from each variable (e.g., example), but cannot separate the impact of each specific example on the performance of the model. 
As our use case is benchmarking, we do not care about, for instance, which example makes the model fail, and hence expect the uniform sampling to be more fitting. 
In practice, the group of all possible prompts is of size $3$, as is available in \helm{}.

\begin{figure}
\centering
\includegraphics[width=\columnwidth]{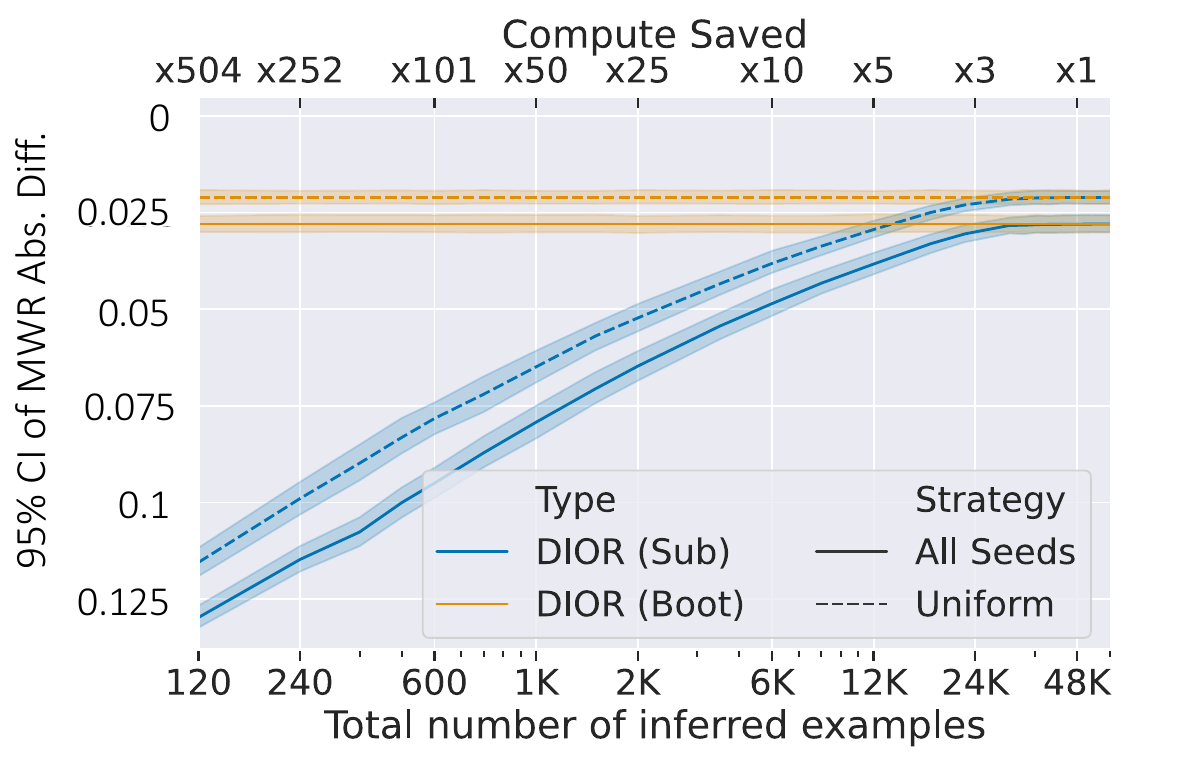}
\caption{\textbf{In-context example selection strategies.} 
The 95\% CI of the MWR difference of bootstrap (\textit{Boot}) and sub-sample (\textit{Sub}) for two choices of In-context example selection (\textit{All Prompts} and \textit{Uniform}). 
It shows that (1) sampling In-context examples uniformly from the pool is superior to using all examples per samples and that (2) more than half of the compute can be saved at no cost to reliability. This analysis discarded the scenarios that did not vary their in-context examples.}
\label{fig:seeds}
\end{figure}

Comparing the two methods in Fig.~\ref{fig:seeds}, we find that the uniform method increases reliability. 
Being limited to only three prompts, we expect this is an underestimation of its true potential. 
Inducting from the prompt-example pairing to the general case, when multiple factors are taken into account, we conclude it is best to sample uniformly, covering as much variability of each factor.

\subsection{Metrics} \label{sec:metric}
Choosing a valid and reliable metric is a complex art, with vast literature. From discussion about metric biases \citep{choshen-abend2018inherent,mathur2020tangled,sulem2018-bleu,peyrard-etal-2021-better}, to metric validation \citep{choshen-abend-2018-automatic,honovich-etal-2022-true,zerva-etal-2022-findings,kocmi-etal-2021-ship,Fabbri2020SummEvalRS}, reference-less metrics \citep{honovich-etal-2021-q2,rei-etal-2022-cometkiwi} and models that evaluate themselves \citep{chia2023instructeval}. 
However, benchmarks that rely on existing datasets (the subscenarios) as their building blocks often adopt their metrics as well. 
Thus, in this analysis, we discuss only the proposed way to convert subscenarios' scores to \helm{}'s score -- MWR. 
This includes two decisions; the grouping of subscenarios into scenarios, where each scenario is weighted similarly (see discussion in~\S\ref{sec:subscenarios}), and the decision to convert the absolute scores per model to a comparative score.

A comparative measure such as win rate provides a preference over models, but can not tell how good a model is at performing a task. 
This is especially useful when preference is easier to collect than an absolute score, as often happens with human evaluation \citep{bojar-etal-2016-findings,choshen-abend-2018-automatic}, or if even direct assessment produces relative scores unintentionally \citep{mathur-etal-2017-sequence,liang2020beyond}. 
Fortunately, this is not our case, where each subscenario provides a score for each example and MWR converts it into pairwise comparisons as a normalization technique.
There are however known and inherent limitations to comparative measures, most famously the impossibility theorems~\citep{arrow1950difficulty}. In this case, introducing a new model to the benchmark changes the scores of existing models \citep{knowles-2021-stability}. 

We analyze if this indeed affects the MWR we currently observe.
Take for example the first two models in \helm{} -- \textit{davinci2} \citep{ouyang2022training} and \textit{Cohere XXL}. 
Those top models switch places when they are compared with or without \textit{Cohere Medium}. This follows from MWR's tendencies. When we introduce models that are just slightly worse in everything than one model, this model sweeps the benchmark collecting all the wins while other models only get some of the wins. Thus, introducing a weaker model changed the rank of two stronger models.

One might consider the example above a rare and extreme case, but actually, this is the expected case. The common practice today is to release several sizes of a new model. Those model variations were trained similarly, and hence tend to have similar strengths, with the larger variant being stronger in every aspect. In App.~\S\ref{ap:examples} we provide a simple numerical example of models changing rank when a new weaker model is introduced.

In essence, one can maliciously raise a model to the top by evaluating numerous models almost equal to their own, but with one wrong sentence in each scenario.
While such intentional gaming is unanticipated, it is a favorable characteristic for a benchmark to improve only if results are better, encouraging innovation through healthy competition.

\section{Efficient Alternative: \flashhelm{}}\label{sec:flashHELM}
In this section, we demonstrate the practical utility of our study, by proposing an efficient variation of the \helm{} benchmark, which we coin as \flashhelm{}. This variation preserves the important information of \helm{}, while reducing computation costs by up to $200$ times.

\begin{figure}[t]
\includegraphics[width=\columnwidth]{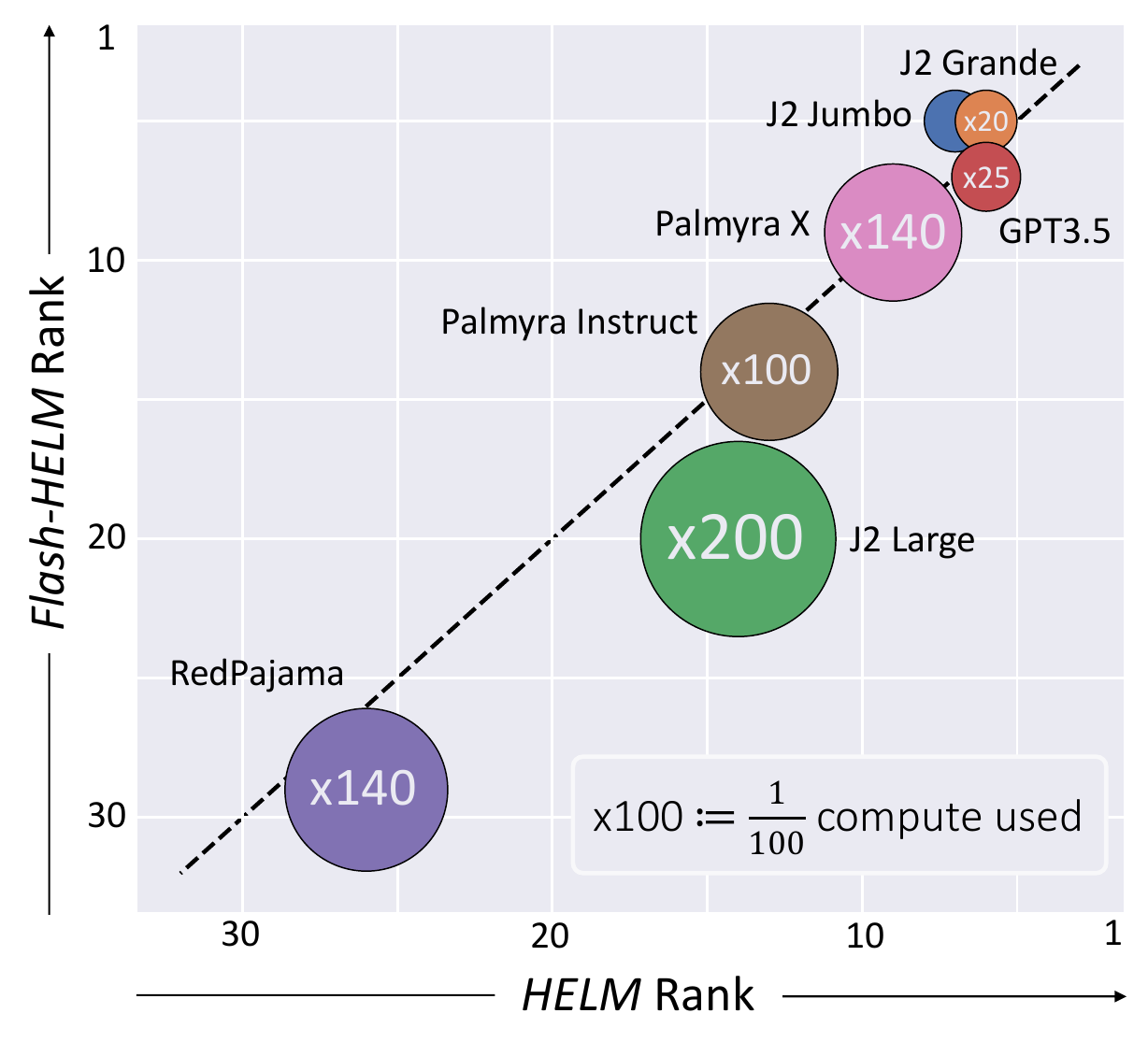}
\caption{\textbf{Efficient evaluation proposal.} \flashhelm{} (\S\ref{sec:flashHELM}) produces similar ranks to \helm{} with a fraction of the compute. Models are Test-set and were not part of the analysis.  Each circle size and numerals represent the reduction in compute usage for evaluation, in comparison to \helm{}'s .
\label{fig:tournament}}
\end{figure}

\paragraph{Objective.}As discussed in \S\ref{sec:reliability}, a well crafted evaluation answers a question. Here we consider the question \textit{``How is model X ranked when compared to other models?''}. Usually, however, the required reliability of the answer varies depending on the model's performance. For example, when a model's ranking falls within the lower range of the benchmark -- say, between positions 25 and 40, the precise ranking might not hold much importance; instead, a broad conclusion that the model is poor should suffice. On the flip side, when a model attains a position in the top $5$ ranks, the specific placement carries more weight.

\paragraph{Approach.}Following this motivation, we propose the following use-case: segmenting the ranking into five ``tiers''; Rank 1, Ranks 2-4, 5-9, 10-19, and Ranks 20 and below. Now, we associate each tier with a designated `desired reliability' level, starting with a low amount of computation for lower ranks and gradually raising it for higher ranks. This allows to evaluate most models tapping into a fraction of the computation. To achieve this improved efficiency while minimally harming reliability, we reduce examples and sample prompts as suggested above.

\begin{figure}[t]
\centering
\includegraphics[width=\columnwidth]{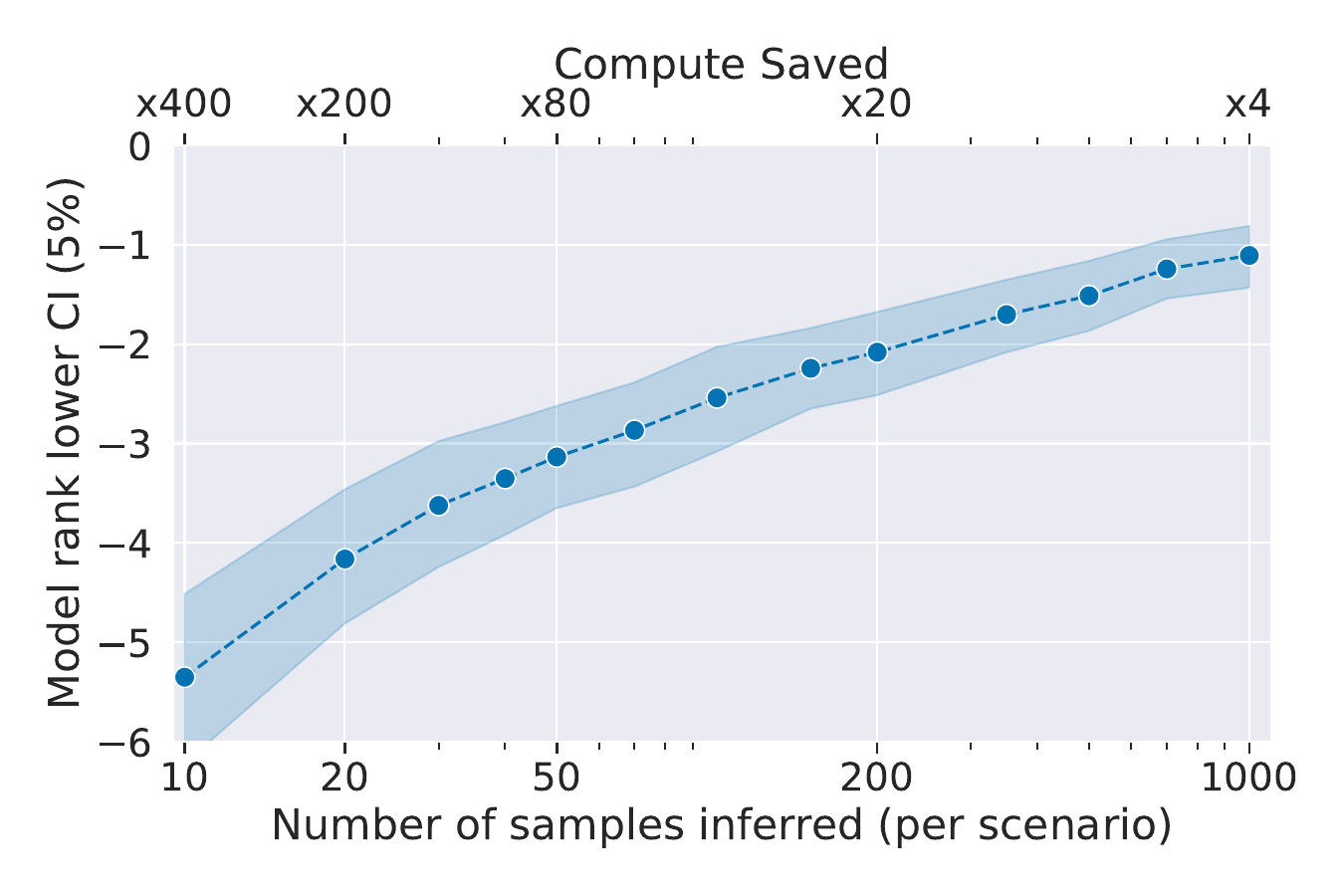}
\caption{Reliable Rank Resolution in \helm{}  \label{fig:locs_res}}
\end{figure}

\paragraph{Algorithm.}
For each tier, 1-4, 5-9, 10-19, and 20 we set the required precision to 1,2,3 and 4 rank resolution respectively. For the top model, we set the precision requirement to be maximal, as identifying the best model bears special importance. Based on that and the data in Figure \ref{fig:locs_res}, we can determine the size of the sub-sample needed in order to evaluate a model in each tier. We denote this relationship as $TierRank(S)$ -- the lower rank of the ``tier'' that is associated with a sub-sample size $S$. Furthermore, we denote: $Rank(M,S)$ - the rank of model $M$ when calculated using a sub-sample size $S$, and $Res(S)$ -- the achieved rank resolution for sub-sample size $S$ (based on Fig.~\ref{fig:locs_res}).
We formulate an efficient `coarse-to-fine' tournament algorithm.
\begin{algorithm}
\caption{Efficient 'coarse-to-fine' tournament}\label{alg:cap}
\begin{algorithmic}
\State $M \gets$ The evaluated model
\For{Sample size $S \in [20, 50, 200, 1000, Max]$}
\State $Rank(M,S) \gets$ Evaluated model $M$ using sub-sample of size $S$.
    \If{$Rank(M,S)-Res(S)\geq TierRank(S)$}
    \State stop;
\EndIf
\EndFor
\State report $Rank(M,S)$.;
\end{algorithmic}
\end{algorithm}

\paragraph{Evaluation.}  We assess the performance of our algorithm using the seven newly-introduced models found in \helm{} v0.2.3 -- models that were not used in our previous analysis. The results are showcased in Figure \ref{fig:tournament}. \flashhelm{} ranks are very close to the full \helm{} ranks and are within the required resolution. These results highlight the algorithm's effectiveness in preserving important ranking information while
achieving a reduction up to a factor of $200$ in computational demands.

\section{Discussion and Conclusion}

Why this sudden focus on reliability when our field largely thrived without it? The shift from single datasets to complex multi-dataset benchmarks, like \helm{}, has changed the evaluation landscape. 
In the past, individual datasets offered great reliability due to a large i.i.d. sample pool spanning all the relevant example space; 
In current benchmarks, on the other hand, the space is constructed of more dimensions such as datasets, prompts, etc. For some of those dimensions (e.g. 3 prompts in \helm{}), the benchmark holds a handful of examples making for a severely low coverage.
Thus, even when the number of examples is sufficient for reliable, and stable insights for some dimensions, this might not be true for others. 
This insufficient coverage narrows the gap with fields like psychology and physiology, which often rely on smaller samples. 
Just as we wouldn't expect meaningful psychological insights from 3 or even 16 human subjects, we shouldn't expect reliable conclusions from just 3 different prompts or 16 datasets without careful design.



Our study shows that by utilizing efficient evaluation methods we can both increase reliability and drastically reduce costs. 
We advocate for the development of more \textit{transparent}, \textit{efficient} and \textit{reliable} evaluation benchmarks and techniques, and by doing so, not only to enhance their effectiveness, but also to make research more accessible across diverse groups, more reproducible and more respectful of environmental concerns. 


\section*{Limitations}
Future work will analyze other benchmarking decisions and other benchmarks. Thus, while the paper's results are sound, they might ignore common unreliable decisions in other benchmarks which were not apparent in this scenario or were left out (such as the choice of prompt templates, choices of non-textual benchmarks, etc.). A decision of special interest is that of efficient inference methods. With many efforts to tackle the tradeoff between performance and computation \citep{chen2023accelerating,choukroun2019low}, future work would wonder if there is a validity and reliability tradeoff as well or if such methods can be used to evaluate models (that do not use them) as well.

In some of the analyses (e.g., tournament) we compare the change with respect to the reported HELM score, as we note throughout the paper, this is but an approximation of the true score each model deserves. Thus, where an efficient method might seem to be deviating in 3 ranks, it might only deviate in 2 (or 4) because the point of reference may itself be wrong. In a sense, reliability compares the change among realizations and solves this problem by not defining which one is the true value.

Here, we considered datasets as sampled and hence similarly informative (except for discussing their correlations). However, it is possible to split datasets into meaningful scenarios. If this would be done for validity reasons one would also want to diversify the scenarios used, perhaps in the space of tasks and domains and skills.

The validity and reliability axes and the claims calling for considering the tradeoffs carefully are general. However, we note that the specific analysis is, specific. It might change in the future, if models characteristics change drastically or improvements make some of the subscenarios redundant. 

Especially prone to that is the rank change, if many similar models are added. In that case, each model would switch more ranks, but as models would still show grouped behaviour and the absolute scores won't change more, we assume the meaningful qualifiers would change as well (for a thousand models, a ±10 in ranks might not be as meaningful as with the current 40).

Another limitation of our work is that we introduce a known and critical aspect in testing, reliability, but evaluate it in an unconventional way. We believe using confidence intervals to be more intuitive and more general and available as anyone running the benchmark already has access to the computation necessary. However, it is more likely that adaptations and improvements would be needed as the traditional statistical study of reliability focuses on variances and such notions.

Our use of bootstrap for experiments (especially with full HELM) has two main limitations. The first is the limitation of bootstrapping in general, while this is the best approximation of the real distribution (e.g., of examples), it is merely an approximation using the sample at hand (HELM's data). 

The second, is that we add an assumption that other decisions could be made that are as valid as the one made by HELM. In other words, we assume there exists a larger distribution from which other choices could have been taken (e.g., instead of considering a summarization task scenario considering paraphrase generation). We do not see that as a strong assumption, as we do not need to explicitly state which distribution that is. If however, the dataset were covering exactly all types of known capabilities or following a theory, that specific choice might not have been a good prospect to test reliability, as it could not change under the circumstances.

Lastly, we would like to note the extension of our work to a single benchmark, although the extent of tasks and models withing this benchmark satisfies our requirements for robust conclusions, adding more benchmarks (e.g.  opencompass~\cite{2023opencompass}) or even specifically running a range of evaluation configurations (using tools such as unitxt~\cite{Bandel2024UnitxtFS}) to further validate our claims remains open for future work.

\clearpage
\bibliography{custom}

\clearpage
\appendix

\section{Mean Win Rate: A Formal Definition}
\label{ap:mwr_def}
For a given set of models $M$, scenarios $CS$ containing subscenarios. Each subscenario provides a single metric to evaluate and score models. For brevity, we identify subscenarios $s{\in }CS$ with their scoring function and define it as $s{\colon }M{\to }\mathbb{R}$: 

\begin{align*} 
&MWR(m)=\\
&\EX_{S\in CS_m}\EX_{m_i\in M\setminus m}\mathds{1}\left(\EX_{s\in{S}}s(m) >  \EX_{s'\in{S}}s'(m_i)\right)
\end{align*} 

Where $\mathds{1}$ is the indicator function. When a subscenario score was not submitted to the benchmark for a specific model $m$ (missing value) it is omitted from $CS$, denoted as $CS_m$.

\section{Examples of Score Sensitivity}\label{ap:examples}
We give several examples of how small changes change the ranking of models without need.

\paragraph{Adding a model.} Take two models with per model scores 10,10,10 and 12,12,8. The second is clearly better. It also gets a better score when the two are compared. However, adding an even worse model 9,9,9 now changes the picture. The win rates of the original models are now 0.5,0.5,1 and 1,1,0 respectively. So on average the scores of the two models are suddenly tied. Adding more such models would improve the first model's ranking more and more, enlarging the difference.

\paragraph{Combining datasets.} Let two models have scores 1,1,0,0 and 0,0,1,1 on 4 datasets respectively. If we call the first two datasets a scenario, we get that one model wins on one scenario and loses on two. This makes the first model suddenly better; choosing the last two datasets would do the contrary.

\paragraph{Reporting partially.} Let three models have average win rates of 0.9,0.9 and 0 and 0.8,0.8,0.8 with many models. If the first model does not report the last 0-winning-rated dataset, then it is considered a better model, with 0.9 win rate on average, while it would be the worse one with 0.6 win rate otherwise.

\begin{figure*}
\includegraphics[width=\textwidth]{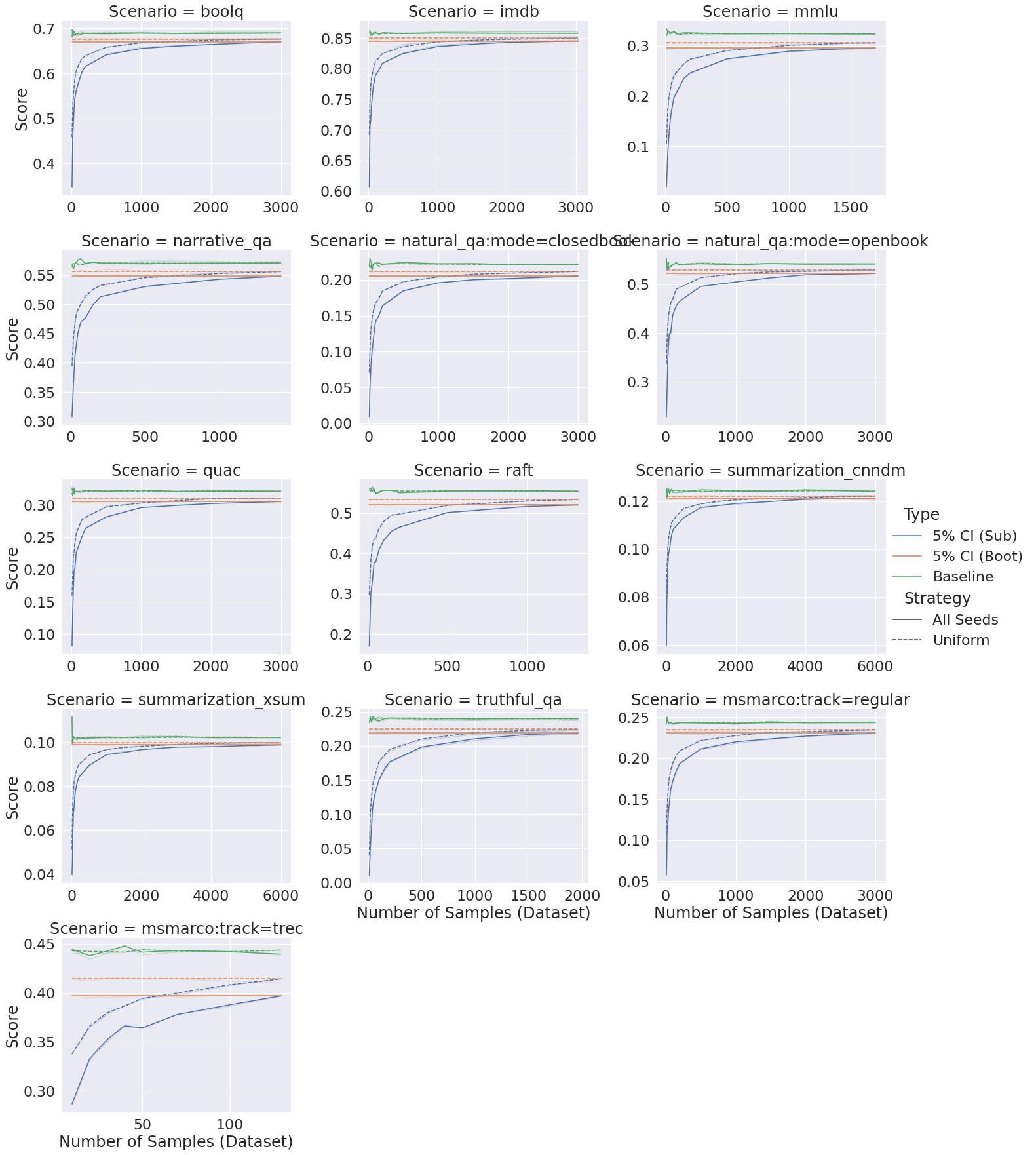}
\caption{\textbf{In-context example selection strategies.} The figure depicts the mean dataset score along with the bootstrap (\textit{Boot}) and sub-sample (\textit{Sub}) 5\% Confidence intervals for two choices of In-context example selection (\textit{All Seeds} and \textit{Uniform}). It shows that (1) sampling In-context examples uniformly from the pool is superior to using all examples per samples and that (2) more than half of the compute can be saved at no cost in score reliability.}
\label{fig:seeds_scores_per_dataset}
\end{figure*}

\section{Objectives per Decision}\label{ap:obj}
\begin{figure*}[phtb]
     \centering
     \begin{subfigure}[b]{0.32\textwidth}
         \centering
         \includegraphics[width=\textwidth]{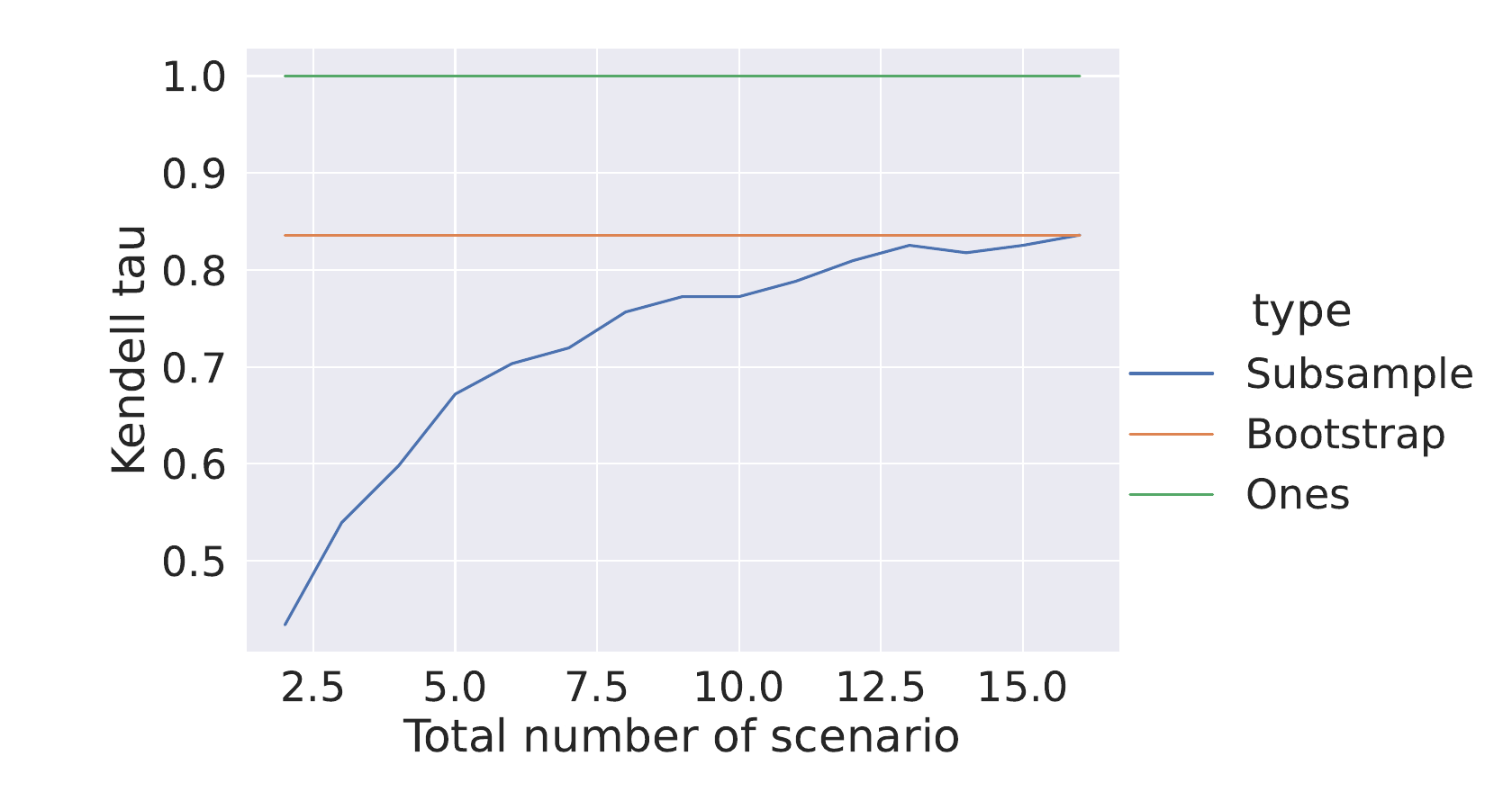}
         \caption{Full Ranking}
         \label{fig:scen_Full_Ranking}
     \end{subfigure}
     \hfill
     \begin{subfigure}[b]{0.32\textwidth}
         \centering
         \includegraphics[width=\textwidth]{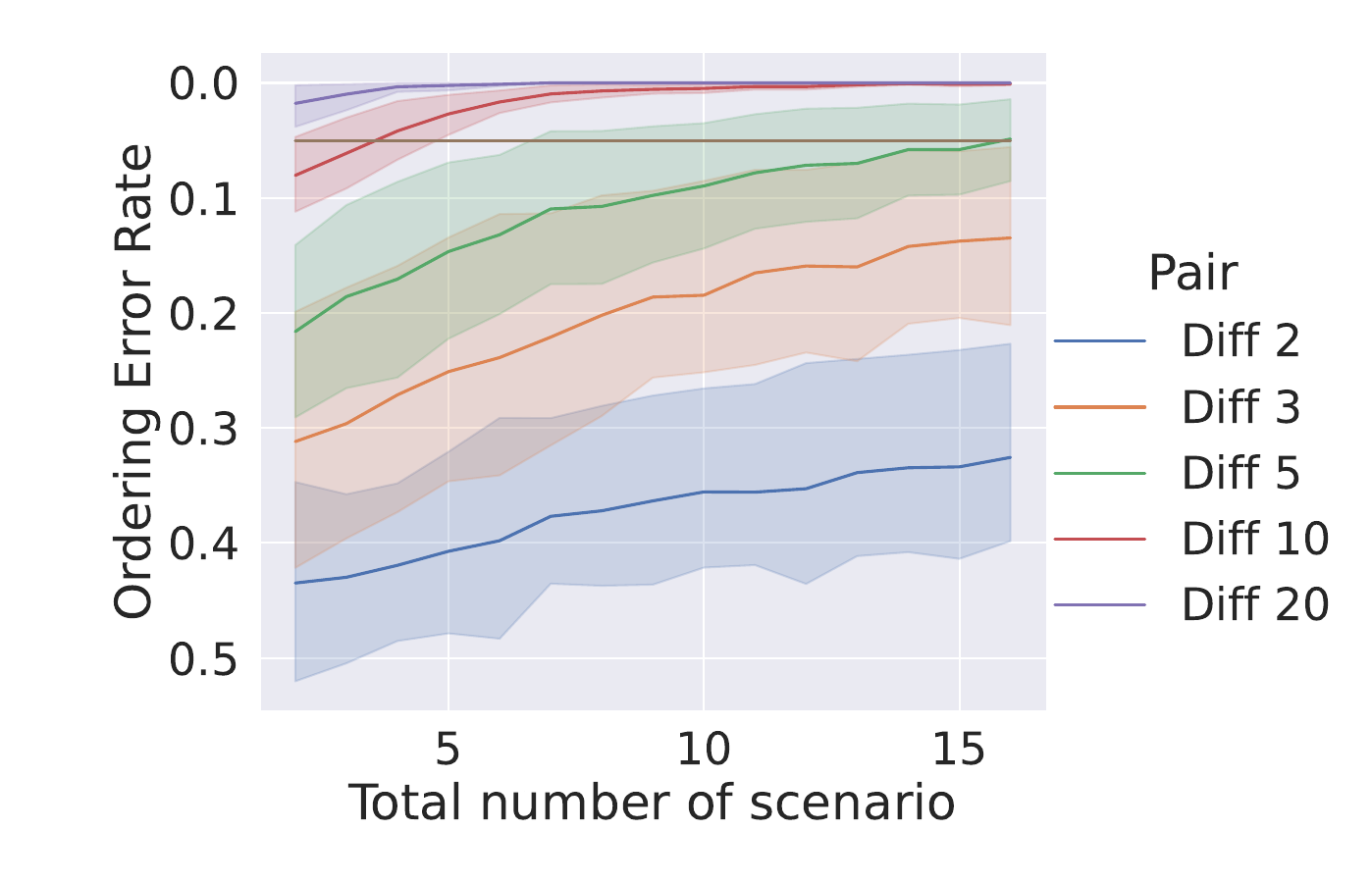}
         \caption{Best Model}
         \label{fig:scen_best_model}
     \end{subfigure}
     \hfill
     \begin{subfigure}[b]{0.32\textwidth}
         \centering
         \includegraphics[width=\textwidth]{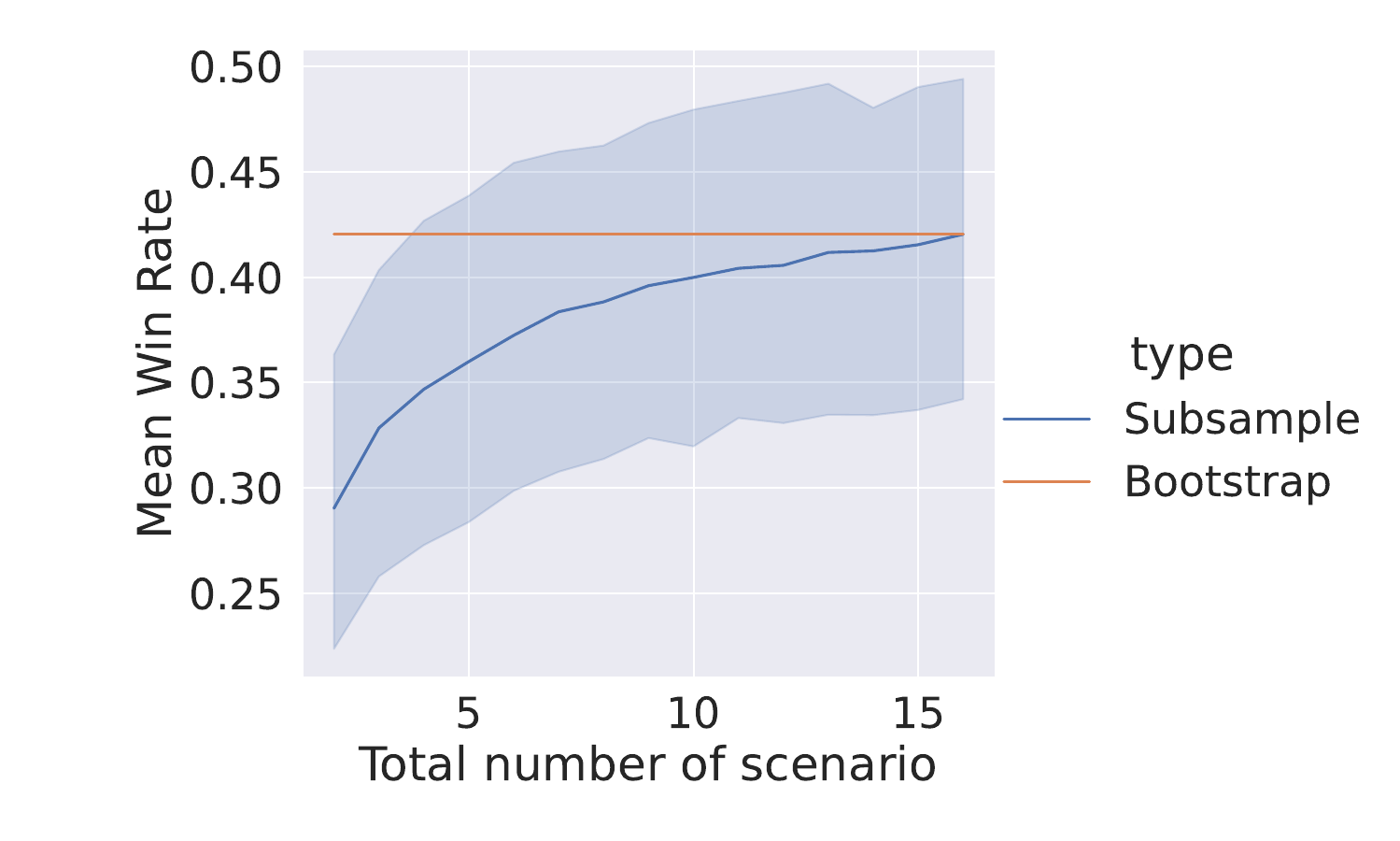}
         \caption{Model Quality}
         \label{fig:scen_model_quality}
     \end{subfigure}
        \caption{Scenarios reliability. Reliability of different amounts of computation for the three objectives and corresponding meta-metrics.
        \label{fig:obj:scen}
        }
\end{figure*}

\begin{figure*}[phtb]
     \centering
     \begin{subfigure}[b]{0.32\textwidth}
         \centering
         \includegraphics[width=\textwidth]{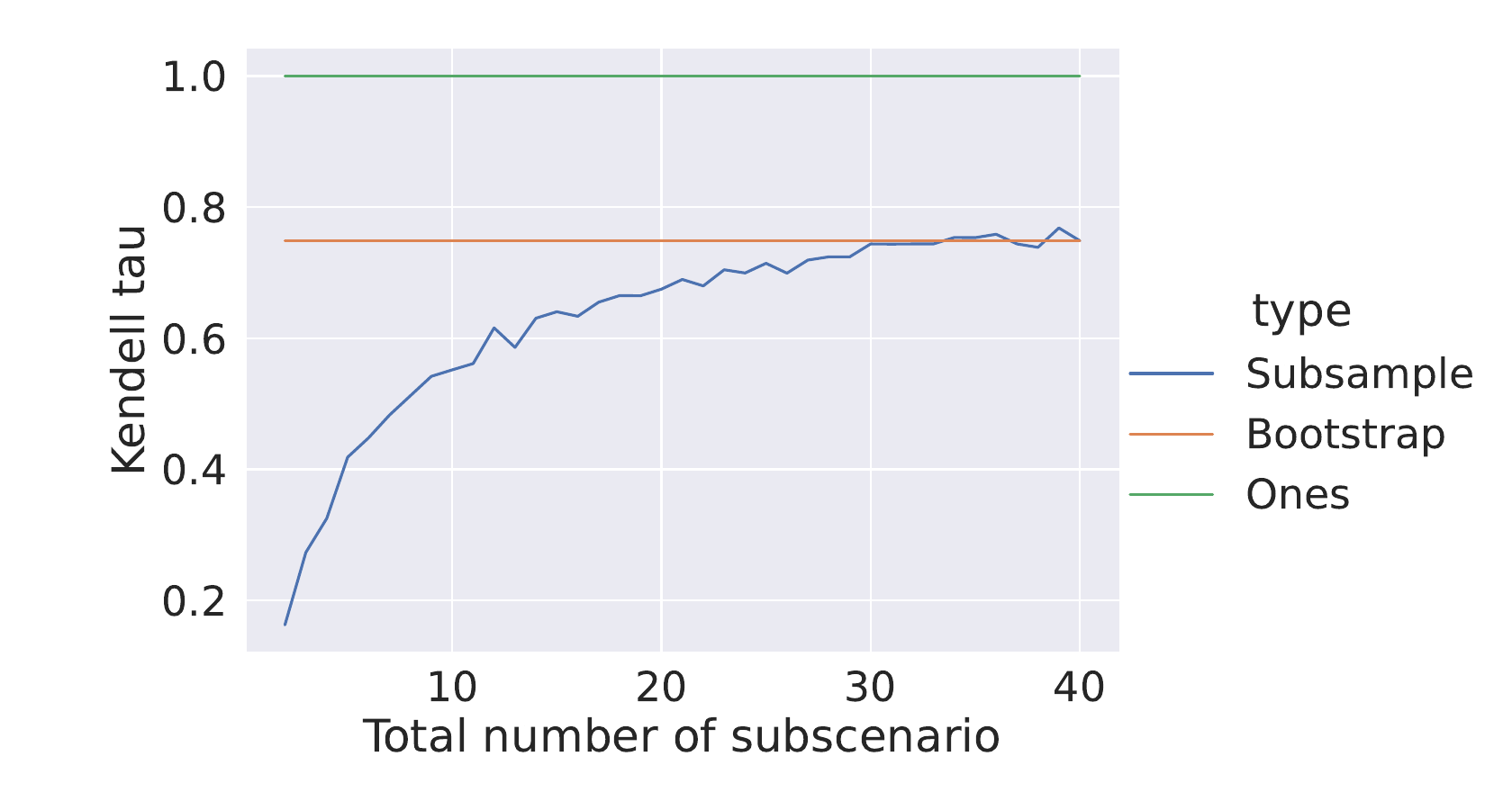}
         \caption{Full Ranking}
         \label{fig:subscen_Full_Ranking}
     \end{subfigure}
     \hfill
     \begin{subfigure}[b]{0.32\textwidth}
         \centering
         \includegraphics[width=\textwidth]{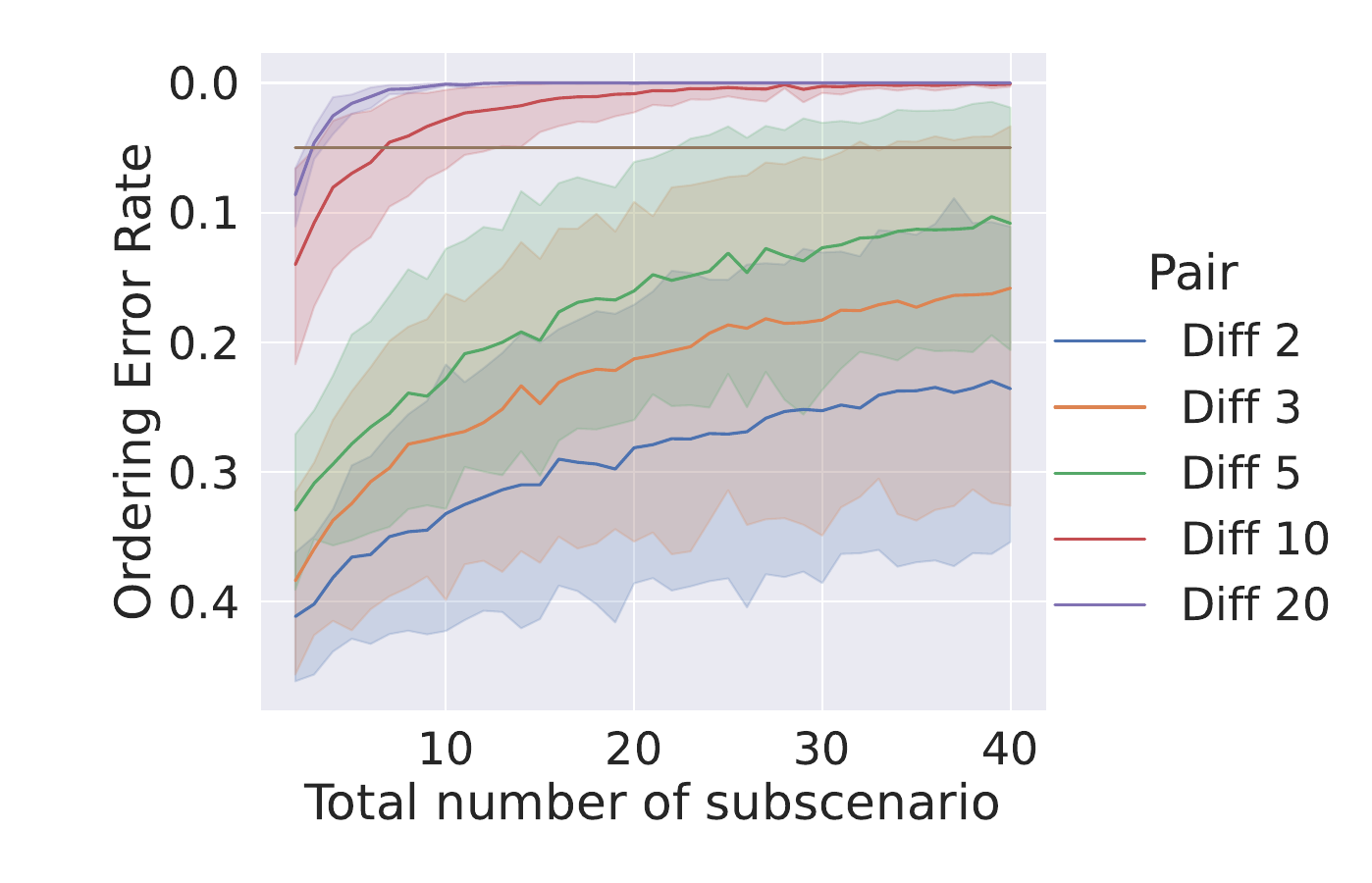}
         \caption{Best Model}
         \label{fig:subscen_best_model}
     \end{subfigure}
     \hfill
     \begin{subfigure}[b]{0.32\textwidth}
         \centering
         \includegraphics[width=\textwidth]{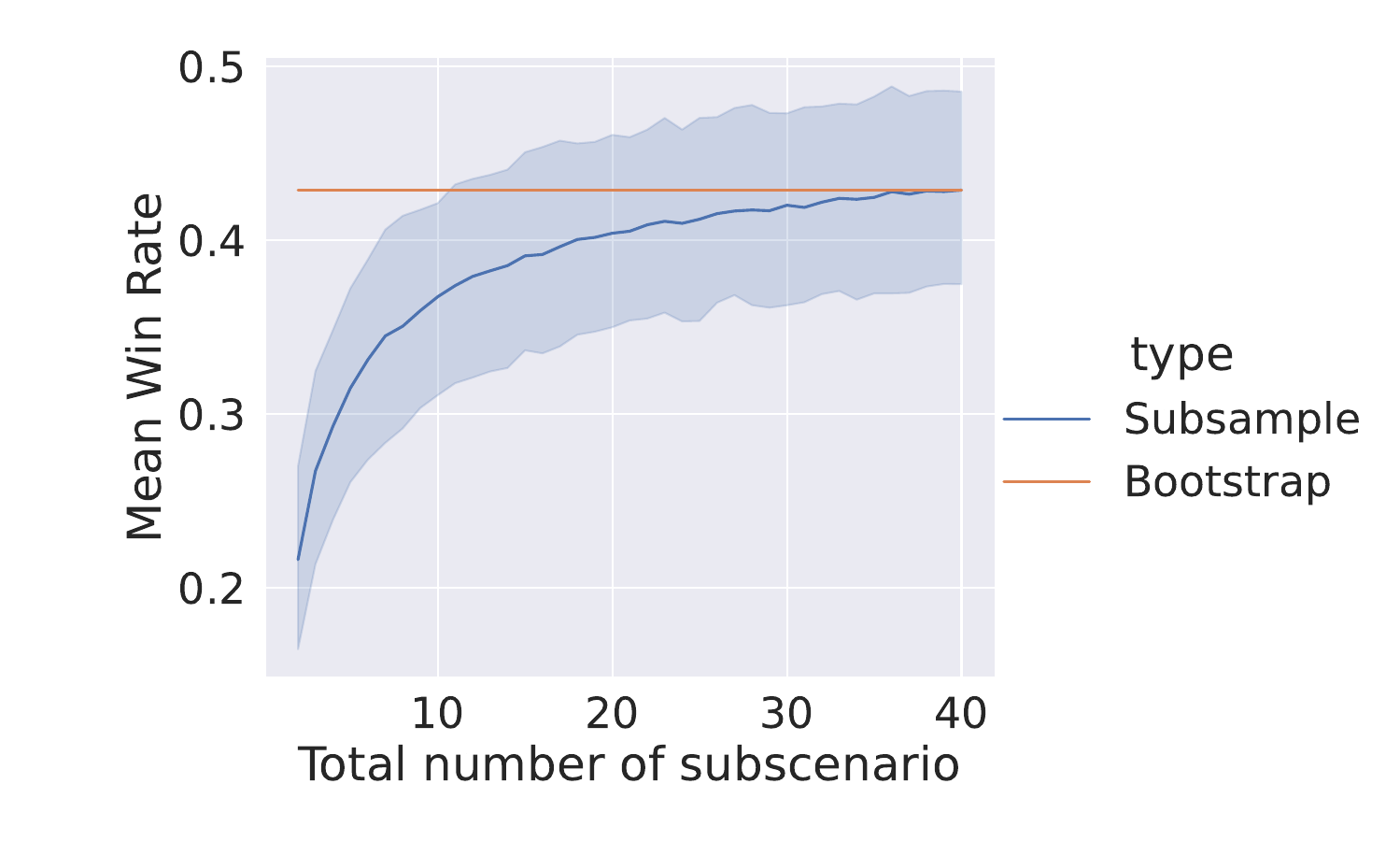}
         \caption{Model Quality}
         \label{fig:subscen_model_quality}
     \end{subfigure}
        \caption{Subscenarios reliability. Reliability of different amounts of computation for the three objectives and corresponding meta-metrics.
        \label{fig:obj:subscen}
        }
\end{figure*}

\begin{figure*}[phtb]
     \centering
     \begin{subfigure}[b]{0.32\textwidth}
         \centering
         \includegraphics[width=\textwidth]{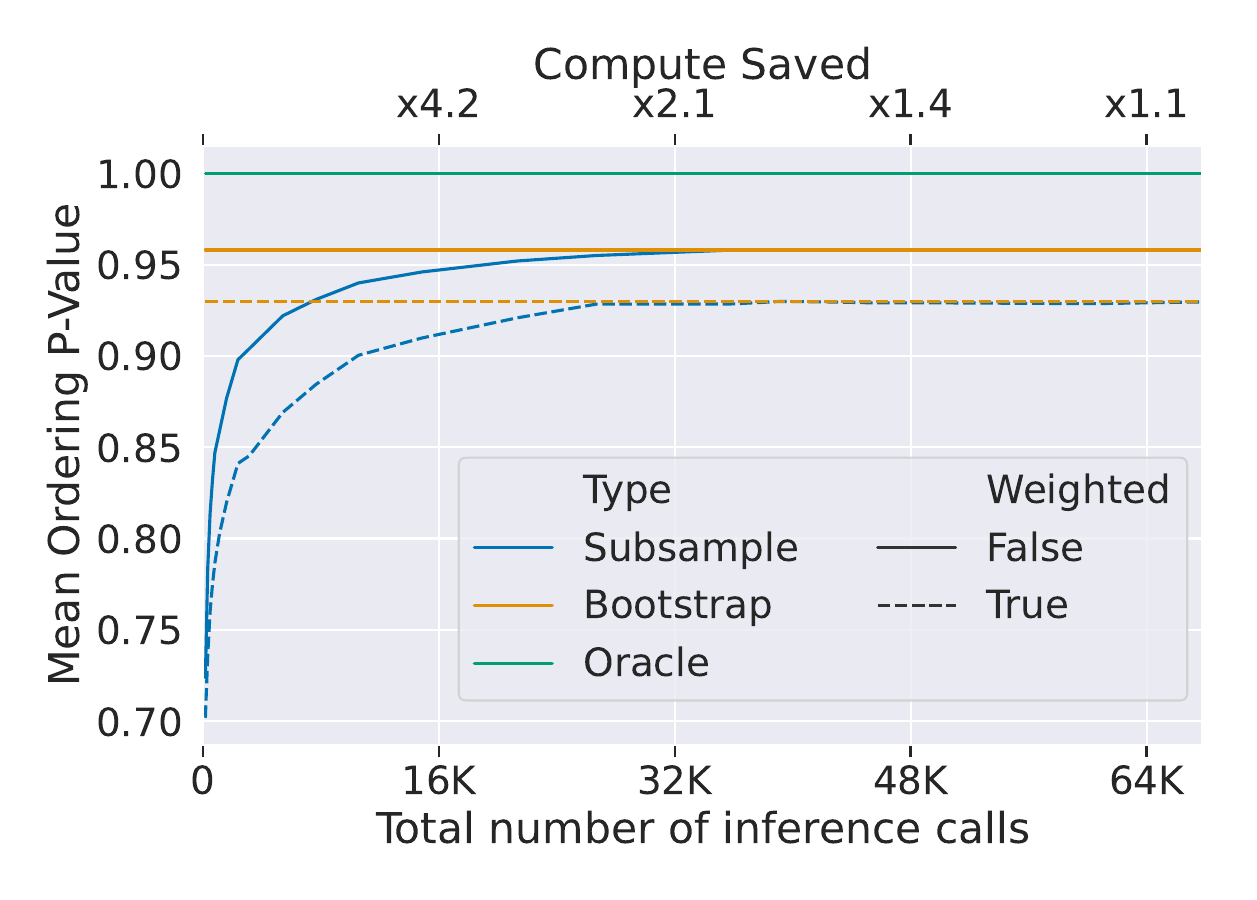}
         \caption{Full Ranking}
         \label{fig:examples_Full_Ranking}
     \end{subfigure}
     \hfill
     \begin{subfigure}[b]{0.32\textwidth}
         \centering
         \includegraphics[width=\textwidth]{graphs/subset_pairsig.pdf}
         \caption{Best Model}
         \label{fig:examples_best_model}
     \end{subfigure}
     \hfill
     \begin{subfigure}[b]{0.32\textwidth}
         \centering
         \includegraphics[width=\textwidth]{graphs/seeds.pdf}
         \caption{Model Quality}
         \label{fig:examples_model_quality}
     \end{subfigure}
        \caption{Examples reliability. Reliability of different amounts of computation for the three objectives and corresponding meta-metrics.
        \label{fig:obj:examples}
        }
\end{figure*}

In this section, we present graphs (\ref{fig:obj:scen},\ref{fig:obj:subscen},\ref{fig:obj:examples}) for decisions and objectives that were left out of the main paper. We provide a triplet of graphs per decision (one for each objective): scenarios in Fig.~\ref{fig:obj:scen}, subscenarios in Fig.~\ref{fig:obj:subscen} and examples in Fig.~\ref{fig:obj:examples}.

\section{Each Dataset as Standalone Benchmark}\label{ap:standalone}
In this section, we report (Fig.~\ref{fig:seeds_scores_per_dataset}) the results separated and without aggregation. We presume this would be helpful both for the use of single standalone benchmarks in the future, and for more elaborate choices when integrating datasets into a new benchmark, such as choosing datasets which provide commendable traits or varying the number of examples shown per dataset in the benchmark.

\section{Full Subscenario Correlations}\label{ap:full_subscenario}
\begin{figure*}
\includegraphics[width=\textwidth]{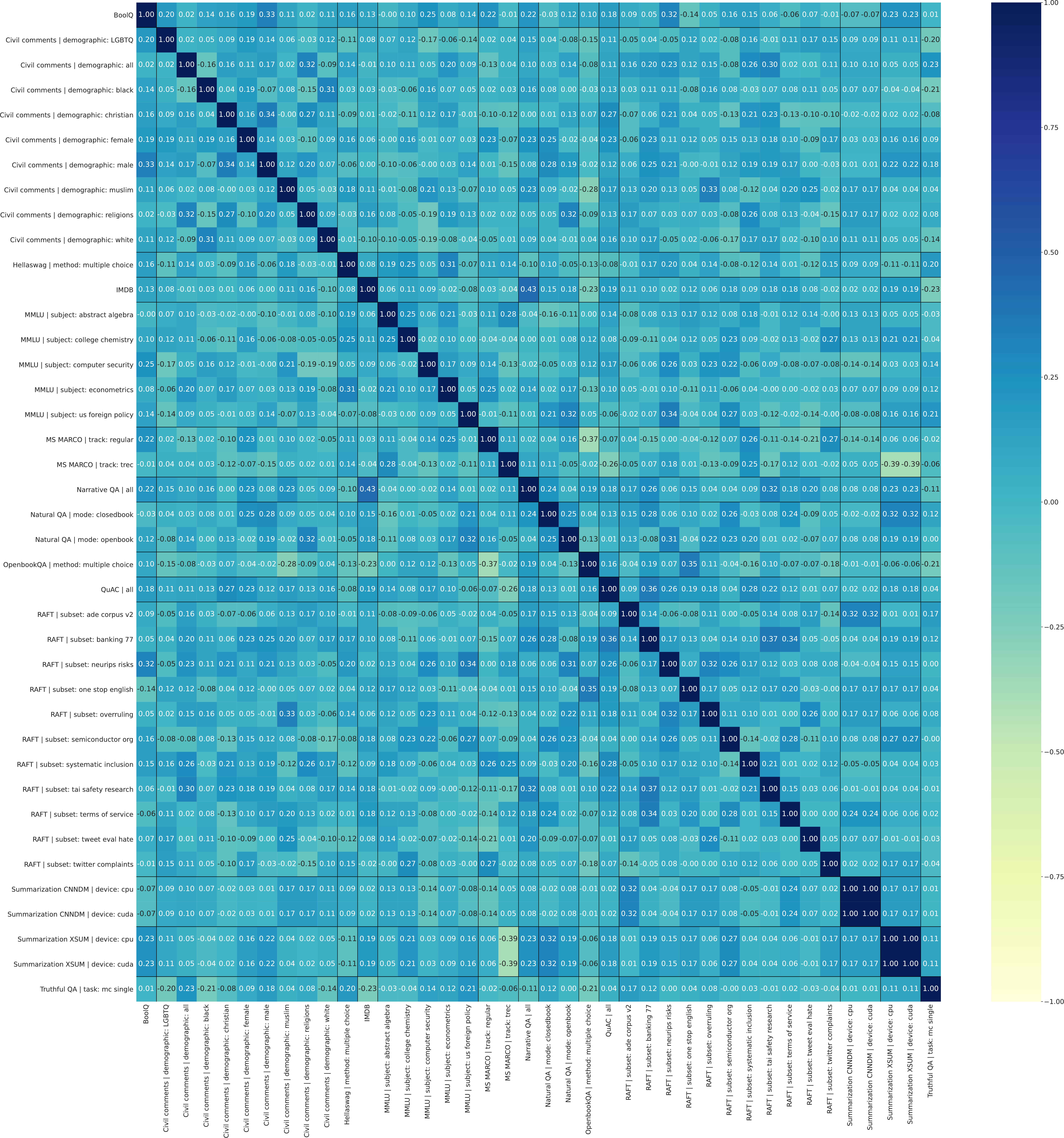}
\caption{\textbf{Subscnerios Ranking Correlations.}  This figure depicts the Kendall $\tau$ correlation matrix between the ranking of models based on the performance in different subscenarios.}
\label{fig:full_subscenario}
\end{figure*}

We provide the full heatmap of correlations between pairs of subscenarios in Fig.~\ref{fig:full_subscenario}, finding little similarity within scenario.

\begin{figure}
\centering
\includegraphics[width=\columnwidth]{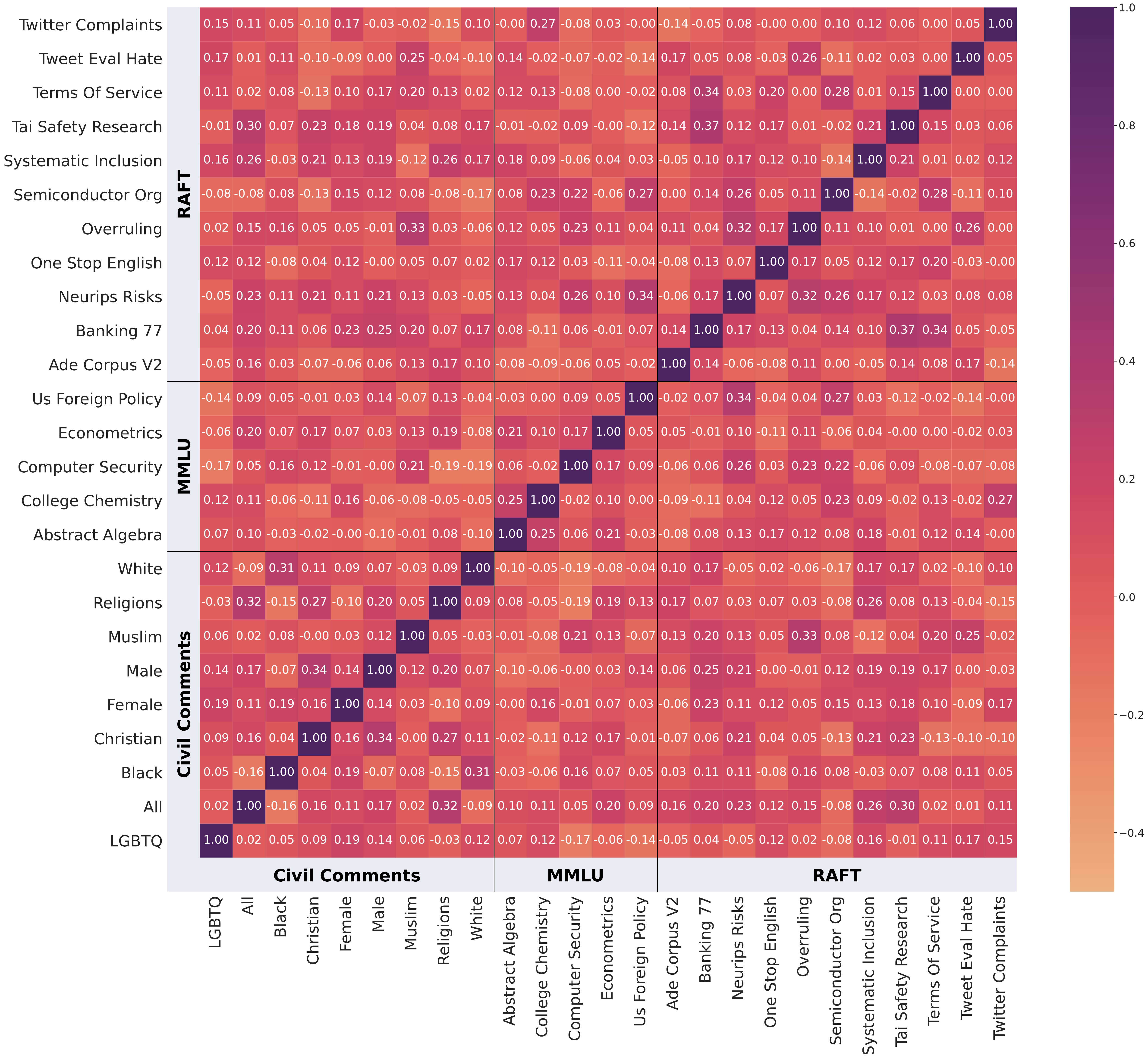}
\caption{\textbf{Subscenerios Ranking Correlations.}  The Kendall $\tau$ correlation matrix between model rankings in each standalone subscenario. Correlations within a scenario are not higher than across scenarios.
\label{fig:subscenario_rank_corr}}
\end{figure}

\section{Scenario Vs Subscenario Aggregation}
\label{ap:scenarios_vs_subscenarios}

\begin{figure}[t]
\centering
\includegraphics[width=\columnwidth]{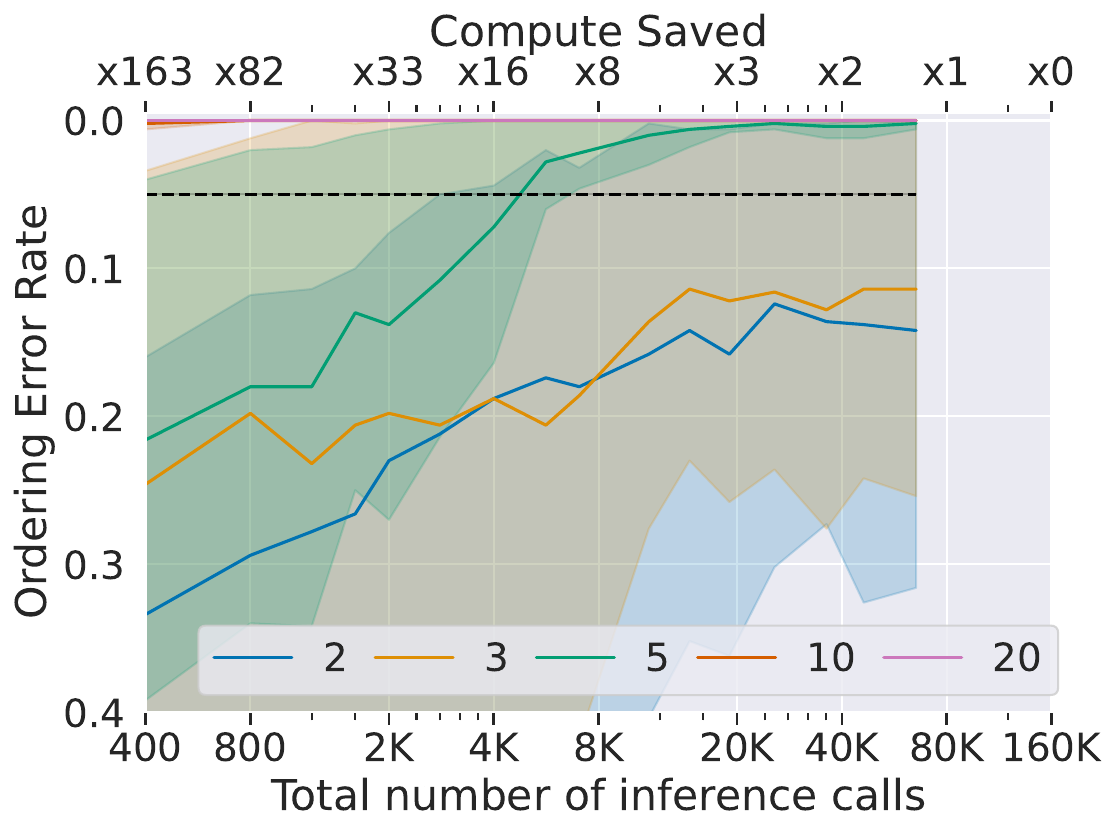}
\caption{\textbf{The probability that models would switch places, MWR over subscenarios} (y-axis) given a different random choice of examples for evaluation (x-axis). Each line corresponds to taking a group of N models and testing if the top and bottom switch places. Results are averaged across $1$K iterations ($95$\% confidence interval in shade) and over the top $5$ models as the top model.}
\end{figure}

\begin{figure}[t]
\centering
\includegraphics[width=\columnwidth]{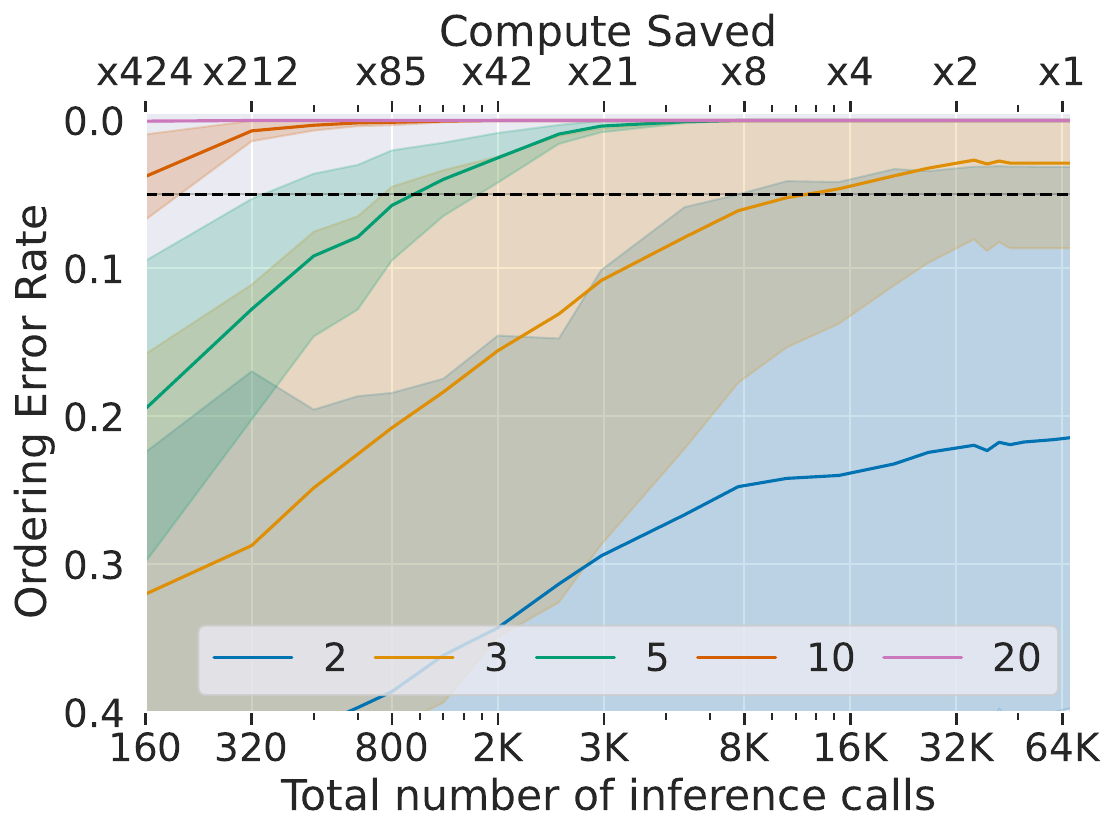}
\caption{\textbf{The probability that models would switch places, MWR over scenarios} (y-axis) given a different random choice of examples for evaluation (x-axis). Each line corresponds to taking a group of N models and testing if the top and bottom switch places. Results are averaged across $1$K iterations ($95$\% confidence interval in shade) and over the top $5$ models as the top model.}
\end{figure}
\end{document}